\documentclass[preprint,review,12pt]{elsarticle}

\usepackage{bbm}
\usepackage{graphicx,subfigure}
\usepackage{amsfonts}
\usepackage{array}
\usepackage{times}
\usepackage{amsmath}
\usepackage{amssymb}
\usepackage{cases}
\usepackage{url}
\usepackage{txfonts}
\usepackage[left=3cm,right=3cm]{geometry}

\makeatletter

\newcommand{\Rmnum}[1]{\expandafter\@slowromancap\romannumeral #1@}
\makeatother

\usepackage{amssymb}

\usepackage{lineno}

\journal{Pattern Recognition}

\begin{document}

\begin{frontmatter}

\title{Pairwise Constraint Propagation: A Survey}

\author[qmul]{Zhenyong Fu}
\ead{z.fu@qmul.ac.uk}
\author[ruc]{Zhiwu Lu}
\ead{zhiwu.lu@gmail.com}
\address[qmul]{School of Electronic Engineering and Computer Science, Queen Mary University of London, Mile End Road, London E1 4NS, United Kingdom}
\address[ruc]{School of Information, Renmin University of China, Beijing 100872, China}

\begin{abstract}
As one of the most important types of (weaker) supervised information in machine learning and pattern recognition, pairwise constraint, which specifies whether a pair of data points occur together, has recently received significant attention, especially the problem of pairwise constraint propagation. At least two reasons account for this trend: the first is that compared to the data label, pairwise constraints are more general and easily to collect, and the second is that since the available pairwise constraints are usually limited, the constraint propagation problem is thus important.

This paper provides an up-to-date critical survey of pairwise constraint propagation research. There are two underlying motivations for us to write this survey paper: the first is to provide an up-to-date review of the existing literature, and the second is to offer some insights into the studies of pairwise constraint propagation. To provide a comprehensive survey, we not only categorize existing propagation techniques but also present detailed descriptions of representative methods within each category.
\end{abstract}


\end{frontmatter}


\section{Introduction}
\label{introduction}

In constrained clustering tasks, people exploit the available prior knowledge to guide the clustering process. Examples of prior information for constrained clustering include relative comparisons \cite{frome2007learning}, pairwise constraints \cite{wagstaff2001constrained}, and cluster sizes \cite{xu2009fast}. Prior knowledge on whether two objects belong to the same cluster or not are expressed respectively in terms of \emph{must-link} constraints and \emph{cannot-link} constraints. Generally, such pairwise constraints, unlike the class labels of data, do not provide explicit class information and are therefore considered a weaker form of supervisory information. In many situations, pairwise constraint relationships between data points are more readily available than the actual class label of the data. Besides the constrained clustering problem \cite{wagstaff2001constrained,klein2002instance,basu2004probabilistic,kulis2009semi}, pairwise constraints have also been widely used for many other machine learning problems such as metric learning \cite{xing2002distance,hoi2006learning,liu2010semi,fu2014local}, and it has been reported that the use of appropriate pairwise constraints can often lead to improved results.

In the constrained clustering research, especially the constrained spectral clustering, people exploit the pairwise constraints for spectral clustering \cite{ng2002spectral,von2007tutorial,shi2000normalized,veksler2008star}, which constructs a new low-dimensional data representation for clustering using the leading eigenvectors of the similarity matrix. Since pairwise constraints specify whether a pair of data points occur together, they provide a source of information about the data relationships, which can be readily used to adjust the similarities between data points for spectral clustering. In fact, constrained spectral clustering has been extensively studied previously. For example, \cite{kamvar2003spectral} (SL) trivially adjusted the similarities between data points to 1 and 0 for must-link and cannot-link constraints, respectively. This method only adjusts the similarities between constrained data points.

However, in general, while it is possible to infer pairwise constraints from domain knowledge or user feedback, in practice, the availability of such constraints is scarce. Hence, one line of research in constrained clustering aims to fully utilize the information inherent in the available pairwise constraints through constraint propagation. It should be noted that the problem of pairwise constraint propagation differs from that of label propagation and is more challenging in the following aspects: (a) unlike class labels for data, pairwise constraints in general do not provide explicit class information; (b) it is in general not possible to infer class label directly from the pairwise constraints which simply state whether a pair of data belong to the same class or not; (c) for a dataset of size $n$, there are potentially $O(n^2)$ pairwise constraints that can be inferred through constraint propagation, while there are only $O(n)$ class labels that need to be inferred for the data in label propagation.

\section{Pairwise constraint propagation methods}

Different from \cite{kamvar2003spectral}, the \emph{pairwise constraint propagation} studies how to propagate the limited available pairwise constraints from the constrained data points to the unconstrained data. For example, \cite{lu2008constrained} (AP) propagated pairwise constraints to other similarities between unconstrained data points using Gaussian process. However, as noted in \cite{lu2008constrained}, this method makes certain assumptions for constraint propagation specially with respect to two-class problems, although a time-consuming heuristic approach for multi-class problems is also discussed. Furthermore, such constraint propagation is also formulated as a semi-definite programming (SDP) problem in \cite{li2008pairwise} (SDP). Although the method is not limited to two-class problems, it incurs extremely large computational cost for solving the SDP problem. In \cite{yu2004segmentation}, the pairwise constraint propagation is also formulated as a constrained optimization problem, but only must-link constraints can be used for optimization.

To overcome these problems in pairwise constraint propagation, an exhaustive and efficient constraint propagation approach ($\mathrm{E^2CP}$) \cite{lu2010constrained,lu2013exhaustive}, which is not limited to two-class problems or using only must-link constraints, has been proposed recently. Specifically, in \cite{lu2010constrained,lu2013exhaustive} the challenging constraint propagation problem is decomposed into a set of independent semi-supervised learning \cite{zhou2004learning,zhu2003semi,lu2009image} subproblems. Through formulating these subproblems uniformly as minimizing a regularized energy functional based on Laplacian regularization \cite{zhu2003semi,lu2011latent}, in \cite{lu2013exhaustive,fu2011symmetric}, it has been shown that the pairwise constraint propagation can be further transformed into solving a continuous-time Lyapunov equation \cite{bartels1972solution} which occurs in many branches of Control Theory such as optimal control and stability analysis \cite{gajic2008lyapunov}. Considering that directly solving the Lyapunov equation scales polynomially to the data size, \cite{lu2010constrained,lu2013exhaustive} further develop an approximate but efficient algorithm based on $k$-nearest neighbor ($k$-NN) graphs using label propagation introduced in \cite{zhou2004learning}. Since the time complexity of the $\mathrm{E^2CP}$ algorithm is quadratic with respective to the data size $N$ and proportional to the total number of all possible pairwise constraints (i.e. $N(N-1)/2)$, it can be considered computationally efficient. As compared to the SDP-based constraint propagation \cite{li2008pairwise} with a time complexity of $O(N^4)$, the $\mathrm{E^2CP}$ algorithm is noted to incur much less time cost. Finally, the resulting exhaustive set of propagated pairwise constraints through the exhaustive and efficient constraint propagation ($\mathrm{E^2CP}$) can be used to adjust the similarity matrix for spectral clustering. Although the $\mathrm{E^2CP}$ constraint propagation and similarity adjustment approaches are proposed in the context of constrained spectral clustering based on pairwise constraint propagation, they can be readily applied to many other machine learning problems initially provided with pairwise constraints.

The methods mentioned previously can only be applied to data within a single modality or presented in a single representation. In reality, there are many datasets with multiple modalities or multiple representations. For example, images found on the Web can either be described by their visual characteristics or by the surrounding texts. Another example is the images from photo sharing websites, such as Flickr. These images can typically be represented using two separate modalities, based respectively on the visual features and the user-provided textual tags. Thus, the problem of pairwise constraint propagation for the multi-modal data has received more and more attention \cite{fu2011multi,lu2012heterogeneous,fu2012modalities,lu2013unified}.

In \cite{fu2011multi}, multiple graphs are constructed for the multi-modal data, with one graph for each of the modalities. Then a random walk process on these graphs is defined. The transition probabilities among the nodes on multiple graphs can be computed using the random walk process. According to such transition probabilities, a series of independent multigraph based constraint propagation subproblems can be formulated. A graph regularization framework is applied to solve these subproblems through a series of quadratic optimizations. Furthermore, it is shown the set of constraint propagation subproblems can be unified and solved as a single quadratic optimization problem and the multi-modal constraint propagation has a closed-form solution.

It should be noted that \cite{fu2011multi} have actually ignored the concept of heterogeneous pairwise constraints or the strategy of heterogeneous constraint propagation. In \cite{fu2011multi}, it is assumed that the constraint settings are consistent among different data modality. In contrast, \cite{lu2012heterogeneous} considers the heterogeneous constraint propagation problem, in which the constraint settings of different data modalities may not be necessarily consistent. \cite{lu2012heterogeneous} is motivated by the homogeneous pairwise constraint propagation method \cite{lu2010constrained}, i.e., the heterogeneous constraint propagation problem can be decomposed into a set of independent semi-supervised learning subproblems which can then be efficiently solved by graph-based label propagation \cite{zhou2004learning}. More importantly, \cite{lu2012heterogeneous} further develop a constrained sparse representation method for graph construction over each modality using the homogeneous pairwise constraints. That is, different from \cite{lu2010constrained} (the homogeneous method), \cite{lu2012heterogeneous} can exploit both heterogeneous and homogeneous pairwise constraints.

In \cite{lu2013unified}, a unified framework for intra-view and inter-view constraint propagation on multi-view data has been proposed. Although both intra-view and inter-view constraint propagation are crucial for multi-view tasks, most previous methods cannot handle them simultaneously. To address this challenging issue, \cite{lu2013unified} propose to decompose these two types of constraint propagation into semi-supervised learning subproblems so that they can be uniformly solved based on the traditional label propagation techniques. To further integrate them into a unified framework, \cite{lu2013unified} utilize the results of intra-view constraint propagation to adjust the similarity matrix of each view and then perform inter-view constraint propagation with the adjusted similarity matrices.

\section{Conclusions}

In summary, pairwise constraint propagation has been proven to be an effective way to exploit the limited number of pairwise constraints. Many pairwise constraint propagation methods have been proposed for both single-view and multi-view data. These methods have been reported to achieve promising results in different challenging tasks in pattern recognition, computer vision, and multimedia content analysis.

\section*{Acknowledgements}

This work was supported by National Natural Science Foundation of China under Grants 61202231 and 61222307, National Key Basic Research Program (973 Program) of China under Grant 2014CB340403, Beijing Natural Science Foundation of China under Grant 4132037, Ph.D. Programs Foundation of MOE of China under Grant 20120001120130, the Fundamental Research Funds for the Central Universities and the Research Funds of Renmin University of China under Grant 14XNLF04, and the IBM Faculty Award.


\end{document}